\documentclass[10pt,twocolumn,letterpaper]{article}

\usepackage[pagenumbers]{cvpr} 

\usepackage{multirow}
\usepackage{microtype}
\usepackage{dblfloatfix}   

\makeatletter
\renewcommand{\paragraph}{\@startsection{paragraph}{4}{\z@}%
  {0.6ex plus 0.2ex minus 0.1ex}{-1em}{\normalfont\normalsize\bfseries}}
\makeatother

\newcommand{\dg}[1]{{\scriptsize\textcolor{red}{#1}}}
\newcommand{\ci}[2]{{\scriptsize[#1, #2]}}

\definecolor{cvprblue}{rgb}{0.21,0.49,0.74}
\usepackage[pagebackref,breaklinks,colorlinks,allcolors=cvprblue]{hyperref}

\def\paperID{HOW-21}
\def\confName{CVPR}
\def\confYear{2026}

\title{Zero-Ablation Overstates Register Content Dependence in DINO Vision Transformers}

\author{Felipe Parodi$^{1}$ \qquad Jordan K. Matelsky$^{1,2}$ \qquad Melanie Segado$^{1}$\\[4pt]
$^{1}$University of Pennsylvania \qquad $^{2}$Johns Hopkins Applied Physics Laboratory\\
{\tt\small \{fparodi, matelsky, msegado\}@upenn.edu}
}

\begin{document}
\maketitle
\thispagestyle{empty}
\renewcommand{\thefootnote}{\fnsymbol{footnote}}
\footnotetext[0]{\textit{Accepted to the \href{https://sites.google.com/view/how-cvpr-workshop/2026-workshop}{HOW Workshop} at CVPR 2026.}}

\begin{abstract}
Zero-ablation---replacing token activations with zero vectors---is widely
used to probe token function in vision transformers.
Register zeroing in DINOv2+registers and DINOv3 produces
large drops (up to $-36.6$\,pp classification, $-30.9$\,pp
segmentation), suggesting registers are functionally indispensable.
However, three replacement controls---mean-substitution, noise-substitution,
and cross-image register-shuffling---preserve performance across
classification, correspondence, and segmentation, remaining within
${\sim}1$\,pp of the unmodified baseline.
Per-patch cosine similarity shows these replacements genuinely perturb
internal representations, while zeroing causes disproportionately large
perturbations, consistent with why it alone degrades tasks.
We conclude that zero-ablation overstates dependence on exact register
content. In the frozen-feature evaluations we test, performance depends
on plausible register-like activations rather than on exact
image-specific values.
Registers nevertheless buffer dense features from
\texttt{[CLS]} dependence and are associated with compressed patch
geometry.
These findings, including the replacement-control results, replicate at ViT-B scale.
\end{abstract}

\begin{figure}[t]
  \centering
  \includegraphics[width=\columnwidth]{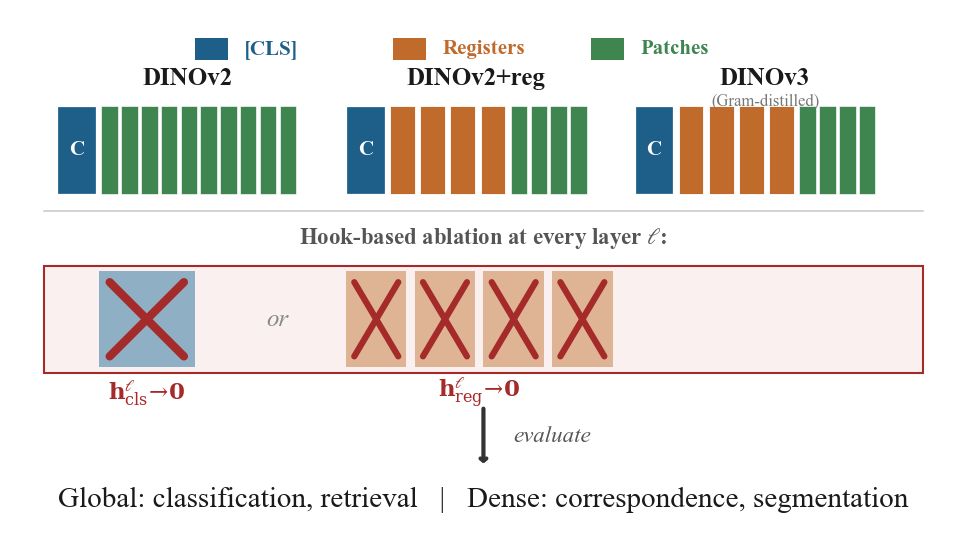}
  \caption{\textbf{Approach overview.}
  We compare three ViT-S/B models with different register configurations. Hook-based ablations zero \texttt{[CLS]} or register hidden states at every block output, and we evaluate on global (classification, retrieval) and dense (correspondence, segmentation) tasks. Three replacement controls test whether observed deficits reflect genuine content dependence or distributional shift from out-of-distribution zero vectors.}
  \label{fig:approach}
\end{figure}

\section{Introduction}
\label{sec:intro}

\begin{figure*}[!t]
  \centering
  \includegraphics[width=0.95\textwidth]{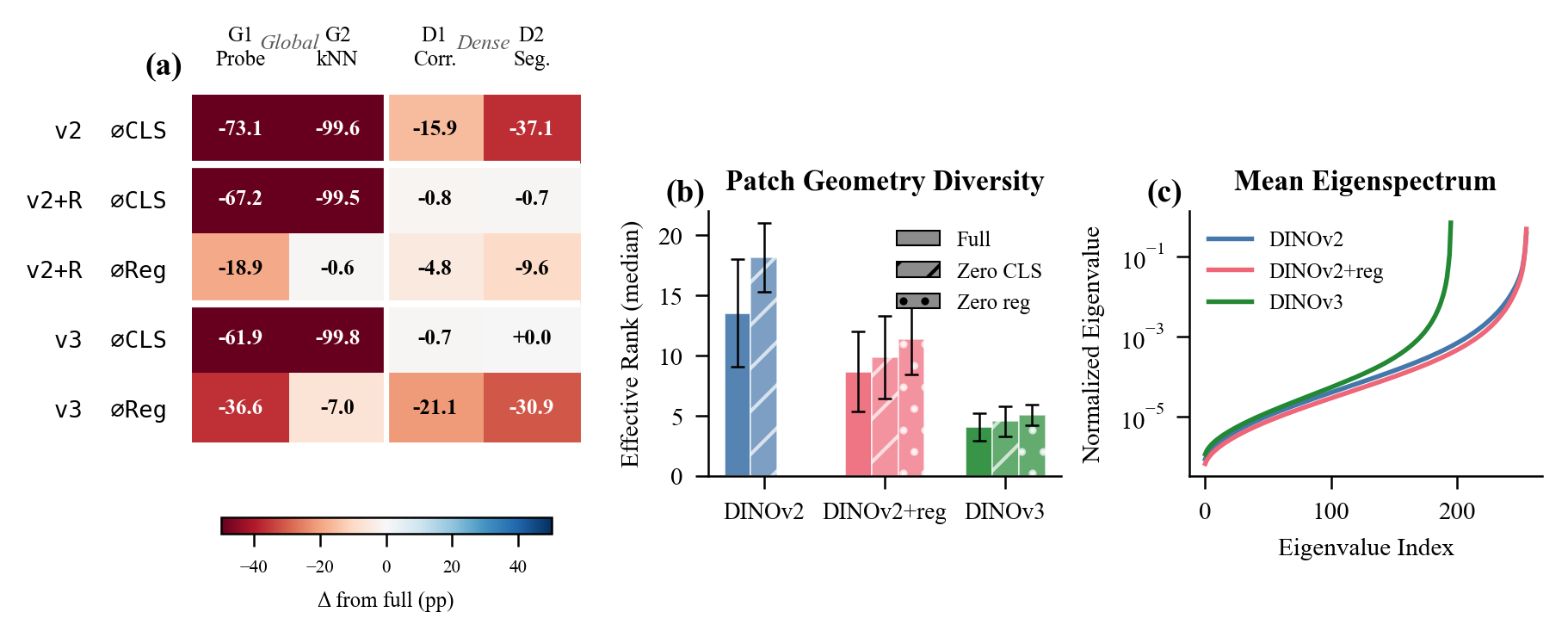}
  \caption{\textbf{Zero-ablation effects and patch geometry.}
  \textbf{(a)}~Task $\times$ Ablation heatmap ($\Delta$\,pp from Full).
  v2\,=\,DINOv2, v2+R\,=\,DINOv2+reg, v3\,=\,DINOv3.
  Zeroing registers produces large drops, but plausible replacement controls
  preserve all tasks (\cref{tab:controls}).
  \textbf{(b)}~Effective rank (median $\pm$ std): registers compress patch
  geometry; DINOv3 exhibits the most compression.
  \textbf{(c)}~Normalized eigenspectrum (log scale, Full condition; eigenvalue index ascending):
  in DINOv3, variance is concentrated in the largest few eigenvalues.
  All models ViT-S.}
  \label{fig:heatmap}
  \label{fig:gram}
\end{figure*}

In neuroscience, lesion studies---selectively damaging neural tissue to infer function---face a well-known confound: damage cascades through interconnected circuits, producing deficits that overstate functional localization~\cite{jonas2017microprocessor}. Zero-ablation in neural networks faces an analogous problem.
Vision transformers (ViTs)~\cite{dosovitskiy2021vit} represent images as a sequence of tokens, including a single \texttt{[CLS]} token that aggregates global information and a grid of \emph{patch tokens} that encode spatially localized features. Self-supervised models such as DINOv2~\cite{oquab2024dinov2} train both token types jointly and now serve as general-purpose backbones across vision domains~\cite{segado2025grounding,parodi2025primateface}, making it important to understand what their internal representations encode. At scale, however, the \texttt{[CLS]} token co-opts patch positions to store global information, producing attention artifacts that degrade dense (patch-level) features~\cite{darcet2024registers}.
\emph{Register tokens}---additional learnable tokens without spatial meaning---absorb this global computation, preventing it from corrupting patch representations~\cite{darcet2024registers}. DINOv3~\cite{simeoni2025dinov3} combines registers with Gram-anchored distillation, achieving state-of-the-art dense features.
Zero-ablation (replacing activations with zero vectors) suggests these registers are functionally indispensable, with zeroing in DINOv3 dropping classification by $36.6$\,pp and segmentation by $30.9$\,pp (\cref{fig:heatmap}).
However, zero vectors are implausible relative to native register activations, and the resulting distributional shift may overstate true content dependence.

We test this directly by applying three replacement controls alongside zeroing: mean-substitution (per-layer dataset means), Gaussian noise matched in mean and variance, and cross-image register shuffling (\cref{fig:approach}). If models depend on register \emph{content}, these replacements should also degrade performance; if they depend only on register \emph{presence}, plausible replacements should suffice.
We apply this framework to DINOv2, DINOv2+registers, and DINOv3 (ViT-S and ViT-B) across classification, retrieval, correspondence, and segmentation.
Jiang~\textit{et~al.}~\cite{jiang2025notrained} showed that \emph{untrained} registers suffice for artifact removal, suggesting register content may not be critical; we test this hypothesis directly in \emph{trained}-register models across downstream tasks.
While recent work has explored representation structure in DINO-family ViTs~\cite{lappe2025decoupling,marouani2026revisiting,fel2026rabbithull}, none has tested whether zero-ablation effects reflect genuine content dependence.

We contribute the following findings.
\textbf{(1)}~Zero-ablation makes registers appear indispensable by injecting out-of-distribution inputs whose effects cascade through all subsequent layers (tens to hundreds of times greater JS divergence than plausible replacements; \cref{fig:rewiring}).
\textbf{(2)}~All three replacement controls preserve performance on classification, correspondence, and segmentation (${\leq}1$\,pp change; \cref{tab:controls}), supporting the conclusion that, in the frozen-feature evaluations we test, models depend on plausible register-like activations rather than image-specific content.
\textbf{(3)}~Per-patch cosine analysis shows these replacements alter internal representations ($0.95$--$0.999$ cosine to full), ruling out the possibility that they leave features unchanged.
\textbf{(4)}~Registers buffer dense features from \texttt{[CLS]} dependence ($37$\,pp segmentation drop without vs.\ ${<}1$\,pp with) and compress patch geometry (effective rank $13.5 \to 4.0$), functioning as structurally expected context channels whose exact per-image content is not uniquely required in the frozen-feature evaluations we test.
These findings, including the replacement-control results, replicate at ViT-B scale (\cref{tab:vitb_controls}).

\begin{figure}[b]
  \centering
  \includegraphics[width=\columnwidth]{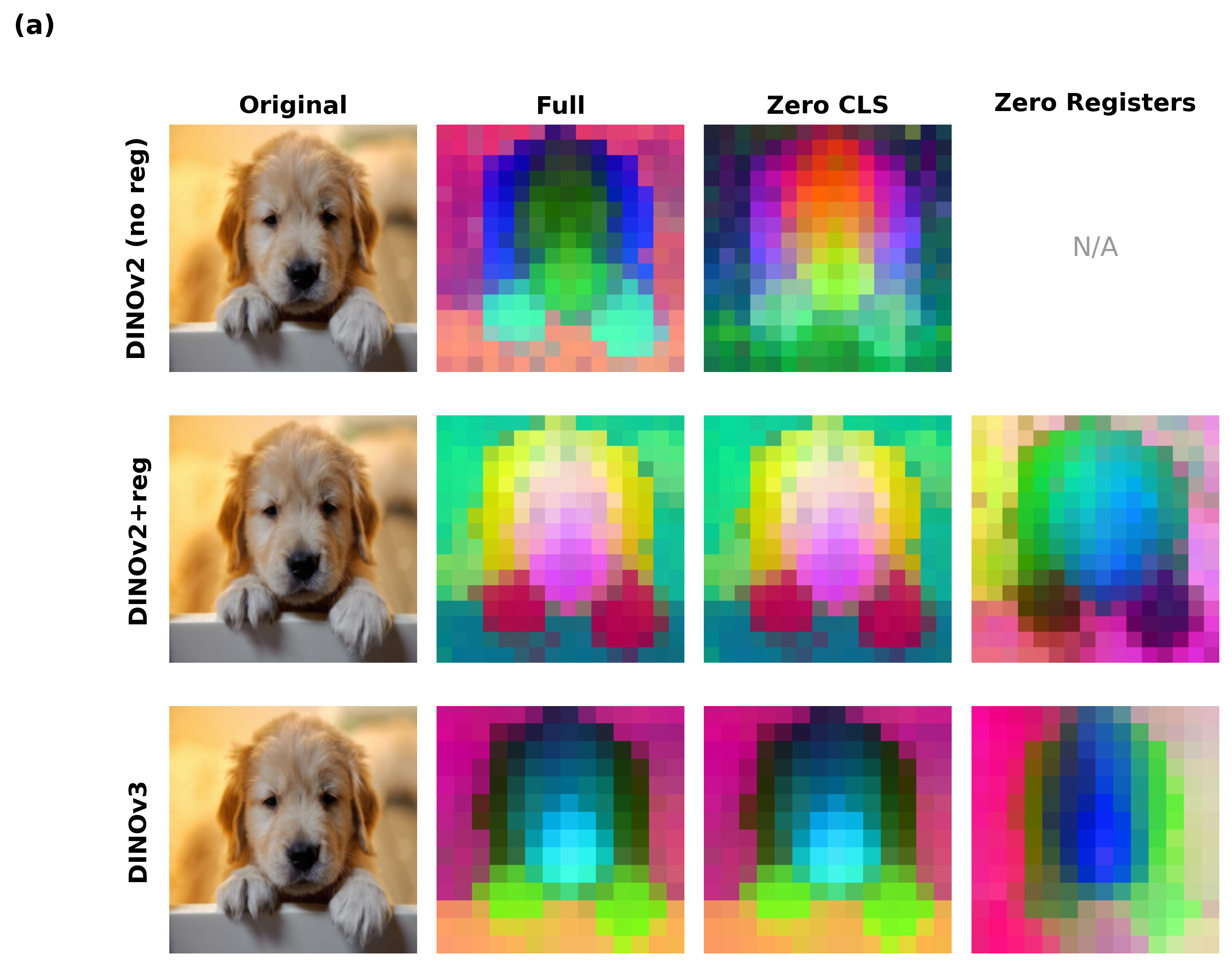}
  \caption{\textbf{PCA projection of patch features under ablation}
  (ViT-S, layer~11, 3-component RGB).
  \emph{Rows}: models. \emph{Columns}: input image, Full (no ablation), Zero CLS, Zero Registers.
  Zero CLS barely alters spatial structure with registers present;
  Zero Registers drastically reorganizes the feature space.}
  \label{fig:pca_ablation}
\end{figure}


\section{Related Work}
\label{sec:related}

\begin{figure}[b]
  \centering
  \includegraphics[width=\columnwidth]{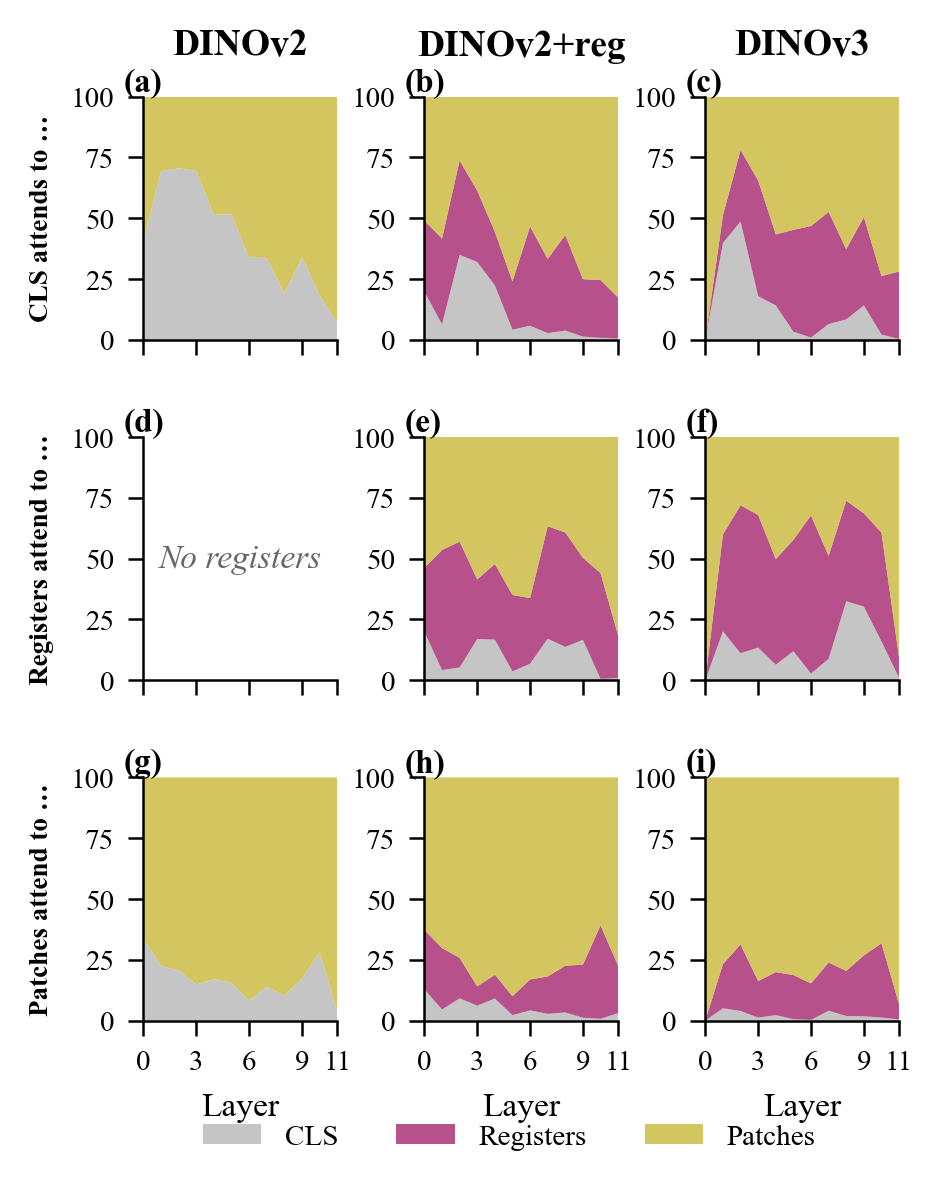}
  \caption{\textbf{Attention flow across all 12 layers (ViT-S, 200 images).}
  Rows: source token type (CLS, registers, patches). Columns: model.
  Stacked areas show what fraction of attention each source directs to CLS (gray), registers (magenta), and patches (olive).
  Register attention builds gradually from mid-layers, yet classification dependence on registers emerges abruptly at layers~10--11 (\cref{fig:gram_dependence}b).}
  \label{fig:attention_flow}
\end{figure}

\paragraph{Register tokens and artifact mitigation.}
Large self-supervised ViTs~\cite{dosovitskiy2021vit,caron2021dino,oquab2024dinov2} produce high-norm artifact tokens at low-informative image regions~\cite{darcet2024registers}; learnable register tokens absorb this computation.
Yellapragada~\textit{et~al.}~\cite{yellapragada2025leveraging} showed register embeddings improve OOD robustness; Shi~\textit{et~al.}~\cite{shi2026morethan} argued registers alone are insufficient, proposing selective patch integration.
Jiang~\textit{et~al.}~\cite{jiang2025notrained} showed untrained registers suffice for artifact removal, suggesting learned register content may not be necessary, but they did not test whether downstream task performance is preserved under content replacement.
Separately, Yan~\textit{et~al.}~\cite{yan2025selfdistilled} showed that post-hoc self-distillation can integrate registers into pretrained ViTs and improve dense representations without full retraining.

\paragraph{Representation structure in DINO-family models.}
Lappe and Giese~\cite{lappe2025decoupling} demonstrated that registers yield local--global feature decoupling, and Marouani~\textit{et~al.}~\cite{marouani2026revisiting} proposed specialized normalization to disentangle CLS and patch processing.
Fel~\textit{et~al.}~\cite{fel2026rabbithull} used sparse autoencoders on DINOv2 to show that downstream tasks recruit distinct concept specializations.
These works characterize the \emph{structure} of token interactions but do not test whether zero-ablation effects on registers reflect true content dependence.

\paragraph{Dense feature degradation.}
Self-supervised ViTs exhibit a tension between global and dense feature quality.
Park~\textit{et~al.}~\cite{park2023whatssl} showed that contrastive learning collapses self-attention, harming dense representations; NeCo~\cite{pariza2025neco} and DVT~\cite{yang2024dvt} address this through patch-ordering and cross-view denoising, respectively.
Beyond register tokens, dense-feature defects can also be mitigated by repairing singular structural artifacts through lightweight fine-tuning~\cite{wang2024sinder}.
DINOv3~\cite{simeoni2025dinov3} regularizes the student's patch Gram matrix against a historical snapshot, preserving second-order geometry during training.
Similar dense-task behavior has been reported in other self-supervised ViTs such as iBOT~\cite{zhou2022ibot}.
How registers interact with these dense features---and whether zero-ablation accurately characterizes that interaction---motivates the present study.

\paragraph{Ablation baselines and the out-of-distribution problem.}
Token ablation is established in NLP~\cite{michel2019sixteen} and adapted for vision efficiency~\cite{bolya2023tome}, but these works ablate to reduce compute, not to probe function.
Zero-ablation conflates removing a component's contribution with pushing activations off-distribution~\cite{hase2021ood}, a concern that parallels the reference-point problem in attribution methods~\cite{sundararajan2017ig}.
In mechanistic interpretability, activation patching---replacing activations with values from other inputs~\cite{geiger2021causal,meng2022locating}---addresses this by keeping activations on-manifold.
Heimersheim and Nanda~\cite{heimersheim2024activation} and Zhang and Nanda~\cite{zhang2024best} showed that ablation-baseline choice can change which components appear causally important in language models, and recommend resample ablation as an in-distribution alternative.
Li and Janson~\cite{li2024optimal} formalized this with optimal ablation, reporting tighter importance estimates than zero or mean baselines.
Our cross-image shuffling control serves an analogous role to resample ablation in this lineage; the mean- and noise-substitution controls test the same in-distribution principle.
Despite the maturity of these tools in NLP interpretability, they have not been applied to register tokens in vision transformers; existing register studies~\cite{darcet2024registers,jiang2025notrained,yan2025selfdistilled} have used zero-ablation without in-distribution controls.


\section{Experimental Setup}
\label{sec:setup}

\paragraph{Models.}
We compare three model families at two scales: DINOv2~\cite{oquab2024dinov2} (no registers), DINOv2+registers~\cite{darcet2024registers} (four register tokens), and DINOv3~\cite{simeoni2025dinov3} (four registers with Gram-anchored distillation), each at ViT-S (21M params, 384-dim) and ViT-B (86M params, 768-dim).
All features are extracted from the final transformer block output (layer~11 of 12) at $224 \times 224$ input resolution.

\paragraph{Intervention framework.}
PyTorch forward hooks replace the relevant token positions (CLS or registers) after every block output, starting from block~1.
This all-layer intervention probes whether the network's forward computation relies on the full token trajectory, not just the final-layer value.
We compare four interventions against the unmodified baseline:
\emph{zero-ablation} replaces activations with $\mathbf{0} \in \mathbb{R}^{d}$;
\emph{mean-substitution} uses per-layer dataset-mean activations calibrated on 5{,}000 ImageNet images (varying $N$ from 100 to 5{,}000 changes accuracy by ${\leq}0.1$\,pp);
\emph{noise-substitution} uses per-layer Gaussian noise matched in mean and variance;
\emph{register shuffling} permutes register activations across images within each batch independently at each layer, preserving real activation structure but breaking image-specific content.

\paragraph{Evaluation tasks.}
We evaluate on four downstream tasks spanning global and dense feature use.
\emph{Classification}: linear probe on a 50{,}000-image ImageNet subset (40K/10K stratified split, seed~42; 1{,}000 classes).
\emph{Retrieval}: $k$NN recall on 2{,}000 ImageNet images with augmented views.
\emph{Correspondence}: nearest-neighbor patch matching on 2{,}000 synthetic pairs and SPair-71k~\cite{min2019spair} (12{,}234 real image pairs).
\emph{Segmentation}: per-patch linear probe on Pascal~VOC~2012 (1{,}464 train / 1{,}449 val images; 21 classes).
Bootstrap 95\% confidence intervals and paired permutation tests (10{,}000 permutations) quantify uncertainty; full training details are in the supplement.

\begin{figure}[t]
  \centering
  \includegraphics[width=\columnwidth]{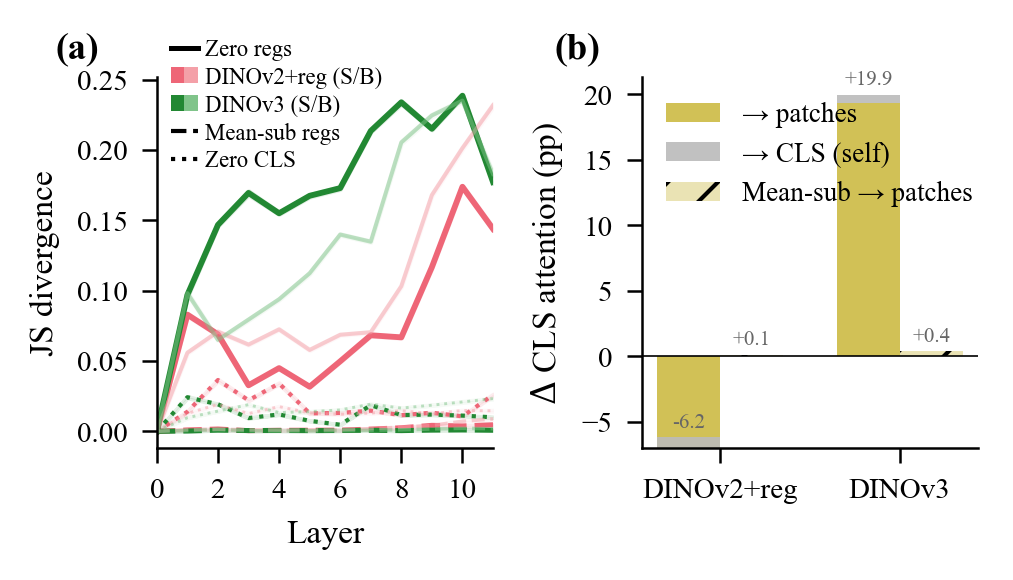}
  \caption{\textbf{Attention patterns under ablation (1{,}000 images).}
  \textbf{(a)}~JS divergence vs.\ layer for ViT-S (dark) and ViT-B (light): register zeroing (solid) causes cascading divergence while mean-substitution (dashed) preserves attention patterns, supporting the distributional-shift interpretation.
  \textbf{(b)}~CLS attention redistribution at last layer when registers are zeroed (ViT-S).}
  \label{fig:rewiring}
\end{figure}


\section{Results}
\label{sec:results}

\subsection{Zero-Ablation Alone Suggests Register Indispensability}
\label{sec:heatmap}

Zeroing register tokens produces large downstream effects (\cref{fig:heatmap}, \cref{tab:heatmap}).
In DINOv2 (no registers), zeroing \texttt{[CLS]} degrades \emph{all} tasks---classification drops to $0.1\%$ (random chance for 1{,}000 classes), correspondence falls $-15.9$\,pp, and
segmentation $-37.1$\,pp---indicating that patches depend on \texttt{[CLS]}.
With registers, dense-task effects become negligible ($-0.8$\,pp and $-0.7$\,pp), buffering patches from \texttt{[CLS]} dependence.
This CLS-zeroing asymmetry ($37.1$\,pp segmentation drop \emph{without}
registers vs.\ $0.7$\,pp \emph{with} them, ${>}50{\times}$; both
$p < 0.001$, paired permutation test; supplement) is consistent with registers absorbing global computation from \texttt{[CLS]}, freeing patches for spatial encoding.

Conversely, zeroing registers in DINOv2+reg reduces CLS classification by $-18.9$\,pp \ci{-20.9}{-17.0} and segmentation by $-9.6$\,pp, while the kNN effect is negligible ($-0.6$\,pp).
DINOv3 shows the same asymmetry but larger in magnitude in both directions: dense features are entirely immune to CLS zeroing
(segmentation $78.5\% \to 78.5\%$), yet register zeroing is far more
destructive---CLS drops $-36.6$\,pp \ci{-38.3}{-34.8} and
segmentation $-30.9$\,pp.
PCA projections (\cref{fig:pca_ablation}) show that Zero~CLS barely alters spatial structure in register models, while Zero~Registers substantially reorganizes it.

Under zero-ablation alone, these results suggest registers are functionally indispensable. However, zero vectors are implausible relative to native register activations, and the resulting distributional shift may overstate true content dependence.

\subsection{Plausible Replacements Preserve All Tasks}
\label{sec:controls}

\begin{table}[t]
\centering
\caption{Register ablation controls (ViT-S, layer~11).
Only zeroing degrades any task; all statistically plausible replacements
preserve performance despite genuinely perturbing internal states
(Patch cos.\ column shows per-patch cosine similarity to full-condition features).}
\label{tab:controls}
\scriptsize
\setlength{\tabcolsep}{2pt}
\begin{tabular}{@{}l cc cc cc cc@{}}
\toprule
 & \multicolumn{2}{c}{CLS (\%)} & \multicolumn{2}{c}{Corr.\ (\%)} & \multicolumn{2}{c}{Seg.\ (\%)} & \multicolumn{2}{c}{Patch cos.} \\
\cmidrule(lr){2-3}\cmidrule(lr){4-5}\cmidrule(lr){6-7}\cmidrule(lr){8-9}
Condition & v2+R & v3 & v2+R & v3 & v2+R & v3 & v2+R & v3 \\
\midrule
Full            & $67.3$ & $62.0$ & $69.1$ & $78.9$ & $71.3$ & $78.5$ & $1.000$ & $1.000$ \\
Zero registers  & $48.4$ & $25.4$ & $64.3$ & $57.8$ & $61.7$ & $47.6$ & $0.605$ & $0.584$ \\
Mean-sub        & $67.0$ & $62.1$ & $68.8$ & $78.8$ & $71.6$ & $78.6$ & $0.959$ & $0.999$ \\
Noise-sub       & $67.0$ & $62.0$ & $68.7$ & $78.7$ & $71.5$ & $78.6$ & $0.955$ & $0.998$ \\
Shuffle         & $67.8$ & $62.0$ & $68.5$ & $78.6$ & $71.2$ & $78.6$ & $0.952$ & $0.998$ \\
\bottomrule
\end{tabular}
\end{table}

\begin{table}[b]
\centering
\caption{Task $\times$ ablation matrix (ViT-S, layer~11).
CLS: probe top-1 (\%). kNN: Recall@1 (\%).
Corr: correspondence (\%). Seg: segmentation mIoU (\%).
$\Delta$: change from Full. Bootstrap 95\% CIs on key deltas in text.}
\label{tab:heatmap}
\scriptsize
\setlength{\tabcolsep}{2pt}
\begin{tabular}{@{}llcccc@{}}
\toprule
& & CLS & kNN & Corr & Seg \\
\midrule
\multirow{2}{*}{DINOv2}
  & Full      & 73.2 & 99.7 & 72.0 & 70.8 \\
  & Zero CLS  & \phantom{0}0.1 \dg{-73.1} & \phantom{0}0.1 \dg{-99.6} & 56.1 \dg{-15.9} & 33.7 \dg{-37.1} \\
\midrule
\multirow{3}{*}{\shortstack[l]{DINOv2\\+reg}}
  & Full      & 67.3 & 99.6 & 69.1 & 71.3 \\
  & Zero CLS  & \phantom{0}0.1 \dg{-67.2} & \phantom{0}0.1 \dg{-99.5} & 68.3 \dg{-0.8} & 70.7 \dg{-0.7} \\
  & Zero Reg  & 48.4 \dg{-18.9} & 99.0 \dg{-0.6} & 64.3 \dg{-4.8} & 61.7 \dg{-9.6} \\
\midrule
\multirow{3}{*}{DINOv3}
  & Full     & 62.0 & 99.8 & 78.9 & 78.5 \\
  & Zero CLS & \phantom{0}0.1 \dg{-61.9} & \phantom{0}0.1 \dg{-99.8} & 78.2 \dg{-0.7} & 78.5 \dg{0.0} \\
  & Zero Reg & 25.4 \dg{-36.6} & 92.9 \dg{-6.9} & 57.8 \dg{-21.1} & 47.6 \dg{-30.9} \\
\bottomrule
\end{tabular}
\end{table}

All three replacement controls preserve performance across classification, correspondence, and segmentation (\cref{tab:controls}):
classification ($67.0$--$67.8\%$ / $62.0\%$),
correspondence ($68.5$--$68.8\%$ / $78.6$--$78.8\%$),
and segmentation ($71.2$--$71.6\%$ / $78.6\%$)---all within ${\sim}1$\,pp of full performance.
Only zeroing causes large drops.
As a negative control, zeroing 4 uniformly random patch tokens (5 seeds) causes ${\leq}1$\,pp drop for ViT-S (${\leq}2.3$\,pp for ViT-B; supplement), indicating that register positions are uniquely vulnerable, not patch tokens in general.

\paragraph{Interpretation.}
In frozen-feature evaluations, register-equipped models depend on the \emph{presence} of plausible register-like activations, not on their specific image-conditioned content.
By contrast, mean-substituting \texttt{[CLS]} yields $0.1\%$ classification (same as zeroing; supplement), indicating that CLS content is image-specific and that the insensitivity to replacement is specific to registers.
That even cross-image shuffling---which breaks all image-specific register trajectories layer by layer---preserves performance is consistent with register geometry being established at training time, requiring only plausible activations at inference.
This conclusion holds for frozen-feature evaluations; whether registers carry task-relevant content under fine-tuning or adaptation remains open.
Register shuffling operates within each batch, so the effective perturbation depends on batch composition.

\paragraph{Perturbation magnitude.}
Per-patch cosine similarity to the full condition (\cref{tab:controls}, last columns) shows that these replacements genuinely alter internal representations---$0.95$--$0.96$ for DINOv2+reg and $0.998$--$0.999$ for DINOv3---while zeroing reduces cosine similarity to $0.60$ and $0.58$, a far larger perturbation consistent with the outsized task degradation.
Jensen--Shannon divergence between full and ablated attention patterns supports this interpretation. Register zeroing yields JS\,$=$\,$0.18$ at the last layer (DINOv3), while mean-substitution yields JS\,$=$\,$0.001$---tens to hundreds of times smaller---consistent with zeroing cascading disruption through all subsequent layers (\cref{fig:rewiring}).

\begin{figure}[b]
  \centering
  \includegraphics[width=\columnwidth]{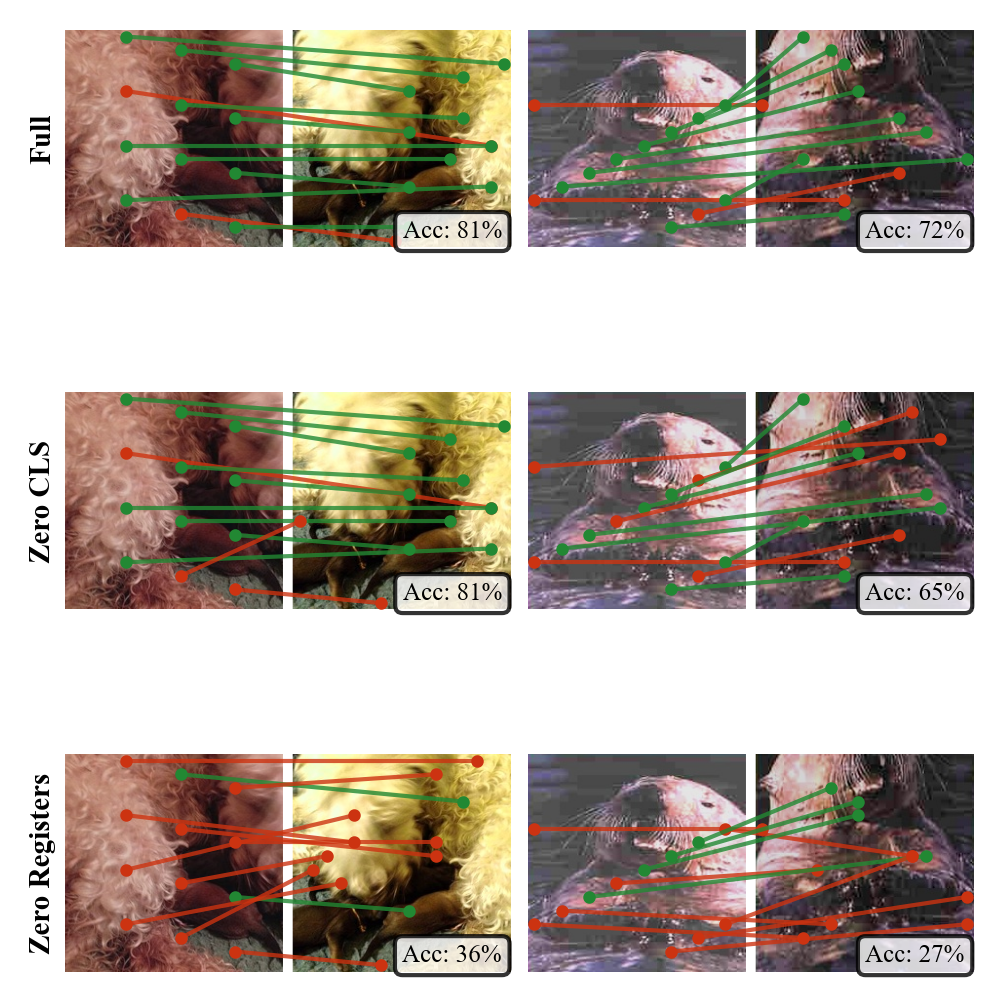}
  \caption{\textbf{Qualitative correspondence under ablation}
  (DINOv3-B, layer~11, tolerance~1).
  Green lines: correct NN matches. Red: incorrect.
  \emph{Top}: Full condition preserves correct correspondences.
  \emph{Middle}: Zero CLS causes minimal disruption (register buffering).
  \emph{Bottom}: Zero Registers collapses correspondence accuracy.}
  \label{fig:correspondence_qual}
\end{figure}

\paragraph{Per-register analysis.}
Initial per-register probes and lesions suggested register specialization (supplement), but the substitution controls show these patterns need not reflect functional dependence: class information is decodable from individual registers, yet models do not appear to require it for the frozen-feature tasks we evaluate.
\cref{fig:correspondence_qual} illustrates the effect qualitatively: register zeroing collapses patch correspondence, while CLS zeroing causes minimal disruption.

\begin{table}[t]
\centering
\caption{SPair-71k PCK@0.1 (\%, ViT-S, layer~11, 12{,}234 pairs).
$\Delta$: change from Full. Bootstrap 95\% CIs in brackets.}
\label{tab:spair}
\scriptsize
\setlength{\tabcolsep}{2pt}
\begin{tabular}{@{}llcc@{}}
\toprule
Model & Ablation & PCK@0.1 & $\Delta$ \\
\midrule
\multirow{2}{*}{DINOv2}
  & Full     & 38.5 \ci{38.0}{39.1} & --- \\
  & Zero CLS & 15.6 \ci{15.3}{16.0} & $-22.9$ \\
\midrule
\multirow{3}{*}{\shortstack[l]{DINOv2\\+reg}}
  & Full     & 35.7 \ci{35.3}{36.2} & --- \\
  & Zero CLS & 35.2 \ci{34.7}{35.7} & $-0.5$ \\
  & Zero Reg & 26.2 \ci{25.7}{26.7} & $-9.5$ \\
\midrule
\multirow{3}{*}{DINOv3}
  & Full     & 31.9 \ci{31.4}{32.3} & --- \\
  & Zero CLS & 32.3 \ci{31.9}{32.8} & $+0.4$ \\
  & Zero Reg & 13.1 \ci{12.8}{13.4} & $-18.8$ \\
\bottomrule
\end{tabular}
\end{table}

\begin{table}[b]
\centering
\caption{ViT-B replacement controls (layer~11).
CLS: probe top-1 (\%). Corr: correspondence (\%).
Seg: segmentation mIoU (\%).
All three controls preserve performance within ${\sim}1$\,pp of Full,
matching the ViT-S pattern (\cref{tab:controls}).}
\label{tab:vitb_controls}
\scriptsize
\setlength{\tabcolsep}{2pt}
\begin{tabular}{@{}l cc cc cc@{}}
\toprule
 & \multicolumn{2}{c}{CLS (\%)} & \multicolumn{2}{c}{Corr.\ (\%)} & \multicolumn{2}{c}{Seg.\ (\%)} \\
\cmidrule(lr){2-3}\cmidrule(lr){4-5}\cmidrule(lr){6-7}
Condition & v2-B+R & v3-B & v2-B+R & v3-B & v2-B+R & v3-B \\
\midrule
Full            & $74.5$ & $73.3$ & $71.2$ & $77.1$ & $72.3$ & $83.4$ \\
Zero registers  & $55.2$ & $36.8$ & $63.3$ & $61.3$ & $64.1$ & $59.6$ \\
Mean-sub        & $74.5$ & $73.4$ & $70.4$ & $76.6$ & $72.7$ & $83.3$ \\
Noise-sub       & $74.3$ & $73.4$ & $70.4$ & $76.5$ & $72.6$ & $83.4$ \\
Shuffle         & $74.7$ & $73.4$ & $70.5$ & $76.8$ & $72.3$ & $83.4$ \\
\bottomrule
\end{tabular}
\end{table}

\begin{figure}[t]
  \centering
  \includegraphics[width=\columnwidth]{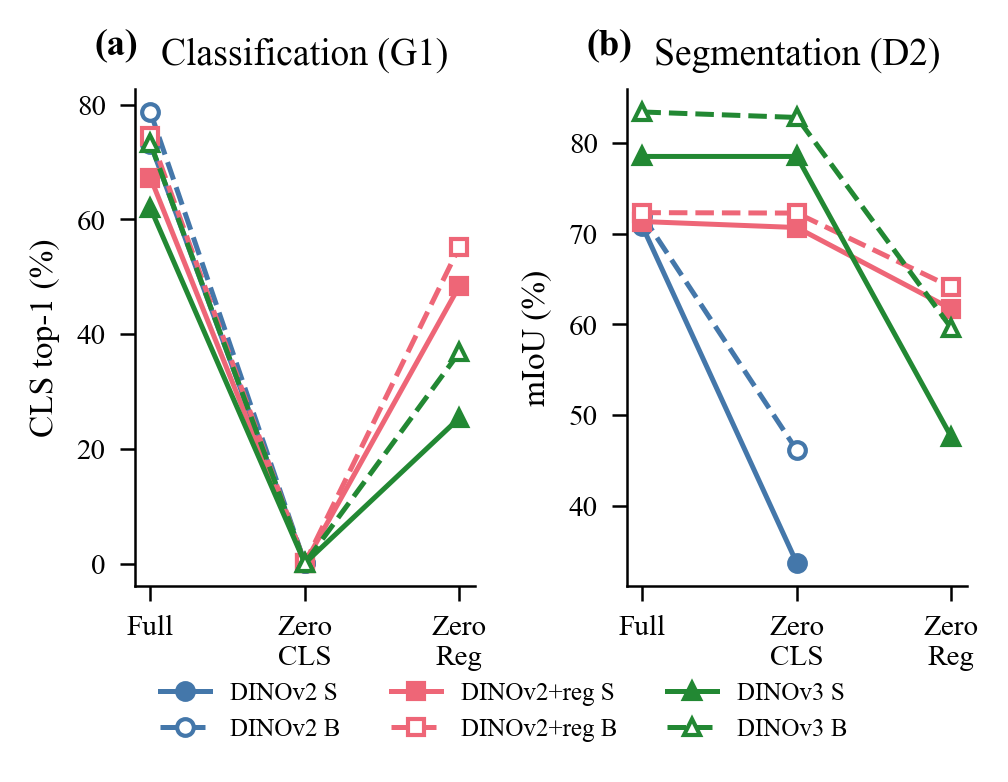}
  \caption{\textbf{Scale comparison: ViT-S vs.\ ViT-B.}
  \textbf{(a)}~Classification and \textbf{(b)}~segmentation under ablation.
  Solid: ViT-S; dashed: ViT-B.
  Ablation delta patterns are consistent across scales
  (\cref{tab:heatmap}).}
  \label{fig:scale}
\end{figure}

\subsection{Structural Effects of Registers}
\label{sec:gram}

If registers offload global computation from patches,
patch representations should occupy fewer effective dimensions.
The patch-to-patch Gram matrix bears this out (\cref{fig:gram}):
at layer~11, median effective rank~\cite{roy2007effective} drops from $13.5 \pm 4.5$ (DINOv2) to $8.7 \pm 3.3$ (DINOv2+reg), a $36\%$ reduction. DINOv3 exhibits the most compressed geometry (effective rank $4.0$, $54\%$ lower than DINOv2+reg).
This compression accompanies improved dense-task performance ($78.5\%$ segmentation vs.\ $70.8\%$ for DINOv2), distinguishing it from the degenerate collapse observed in contrastive learning failures~\cite{park2023whatssl}.
Because DINOv3 simultaneously differs in patch size, positional encoding (RoPE), and distillation recipe, we cannot attribute this additional compression solely to Gram anchoring; all three factors may contribute.
This compression is already present at intermediate layers---DINOv3's effective rank is $6.4$ at layer~6 vs.\ $32.3$ for DINOv2---yet register dependence for classification emerges only at the final layers (\cref{fig:gram_dependence}).
Attention routing and functional dependence are dissociated. CLS$\to$register attention share builds gradually from mid-layers (${\sim}20\%$ by layer~6 in DINOv2+reg, ramping to $29\%$ at layer~11 in DINOv3; \cref{fig:attention_flow}), but classification accuracy under register zeroing remains near-random until layers~10--11, suggesting that the functional coupling is late-stage even though the structural routing develops early.

\paragraph{Attention routing.}
CLS attention corroborates this structural role: DINOv2+reg directs $17.9\%$ and DINOv3 $29.1\%$ of last-layer CLS attention to registers (supplement), paralleling the increasing register dependence observed under zero-ablation.

\paragraph{Real-image generalization.}
The asymmetry extends beyond ImageNet to real-image correspondence.
On SPair-71k~\cite{min2019spair} (12{,}234 pairs, 18 categories; \cref{tab:spair}), CLS zeroing degrades correspondence by $-22.9$\,pp in DINOv2 but only $-0.5$/$+0.4$\,pp with registers, while register zeroing produces the reverse ($-9.5$/$-18.8$\,pp).
The same pattern holds at ViT-B ($-18.8$\,pp DINOv3-B vs.\ $-12.3$\,pp DINOv2-B+reg under register zeroing; supplement).

\subsection{Scale Validation}
\label{sec:scale}

All findings replicate at ViT-B scale (\cref{tab:vitb_controls}, \cref{fig:scale}).
The zero-ablation asymmetry is preserved: DINOv3-B loses $-36.5$\,pp classification and $-23.8$\,pp segmentation under register zeroing, while CLS zeroing has negligible dense-task effects (${\leq}0.6$\,pp; supplement).
The replacement-control result also replicates: all three controls preserve classification ($74.3$--$74.7\%$ vs.\ $74.5\%$ full), correspondence ($70.4$--$70.5\%$ vs.\ $71.2\%$), and segmentation ($72.3$--$72.7\%$ vs.\ $72.3\%$) for DINOv2-B+reg, with even tighter margins for DINOv3-B (${\leq}0.4$\,pp across all tasks).
The Gram compression pattern also holds: effective rank drops $20.1 \to 13.9 \to 6.6$ at ViT-B ($3.0{\times}$ compression from DINOv2-B to DINOv3-B, vs.\ $3.4{\times}$ at ViT-S).
The attention-pattern asymmetry is consistent at ViT-B: register zeroing yields last-layer JS divergence of $0.232$ (DINOv2-B+reg) and $0.183$ (DINOv3-B), while mean-substitution yields $0.009$ and $0.002$.

\section{Discussion}
\label{sec:discussion}

\begin{figure}[t]
  \centering
  \includegraphics[width=\columnwidth]{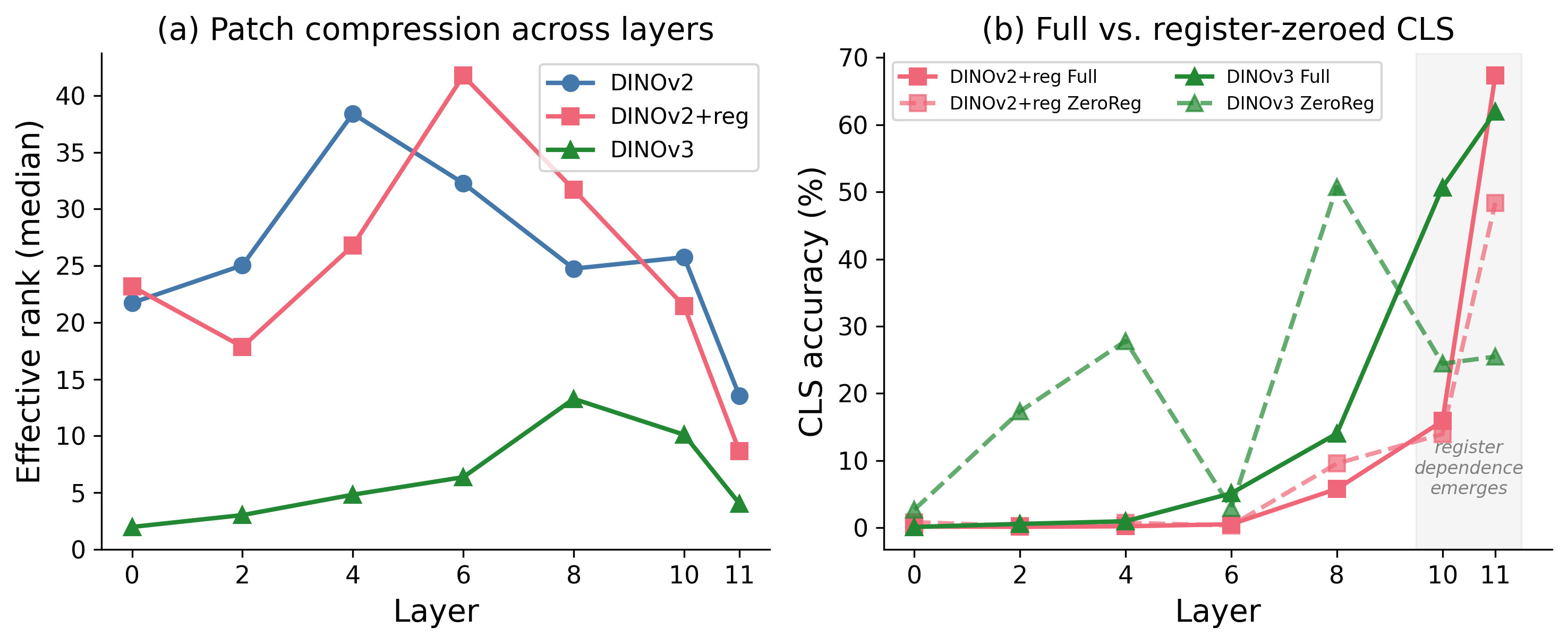}
  \caption{\textbf{Patch compression and register dependence across layers.}
  \textbf{(a)}~Effective rank (Full condition) decreases across layers; DINOv3 is already compressed by layer~6.
  \textbf{(b)}~CLS accuracy under Full (solid) vs.\ register-zeroed (dashed):
  dependence is layer-specific, emerging only at layers~10--11. ViT-S models.}
  \label{fig:gram_dependence}
\end{figure}

Zeroing register tokens injects vectors that occupy a distinct region of activation space, and the resulting distributional shift cascades through all subsequent layers. Three independent replacement controls show that, in frozen-feature evaluations, performance depends on the presence of plausible register activations rather than on exact image-specific values.
The network's computation has reorganized around register slots, and registers buffer dense features from \texttt{[CLS]} dependence.

Several limitations bound these conclusions.
Our all-layer intervention probes whether the network relies on the full register trajectory; single-layer or late-layer-only ablations may reveal layer-specific dependencies not captured here.
The insensitivity to register replacement is established for standard frozen-feature evaluations (classification, correspondence, segmentation); whether it holds for tasks that may require fine-grained register content (few-shot adaptation, generation) or for non-register token types in other architectures (e.g., \texttt{[CLS]} in DeiT~\cite{touvron2021deit} or MAE~\cite{he2022mae}) remains open.
Our between-model comparisons are correlational; disentangling Gram anchoring from co-occurring architectural differences (patch size, RoPE, distillation) requires controlled ablation of the training objective itself.

The parallel to neuroscience is direct. Lesion-induced deficits can reflect network disruption rather than localized function~\cite{jonas2017microprocessor}, and zero-ablation overstates dependence whenever the ablated component occupies a distinct region of activation space.
Cross-image shuffling provides the strongest test of content dependence, while mean-substitution is simplest to implement; we recommend either as a standard complement to zero-ablation.

\section{Conclusion}
\label{sec:conclusion}

In the frozen-feature evaluations we test, register tokens in DINO-family ViTs act as structurally expected context channels. Performance depends on plausible register activations rather than on exact image-specific values.
Zero-ablation overstates register dependence because it injects uniquely destructive out-of-distribution vectors, not because registers carry content indispensable for these tasks.
Registers nonetheless play a genuine structural role, buffering dense features from \texttt{[CLS]} dependence and compressing patch geometry.
These findings replicate at ViT-B scale.

{
    \small
    \bibliographystyle{ieeenat_fullname_unsorted}
    \bibliography{main}
}

\clearpage
\maketitlesupplementary


\renewcommand{\topfraction}{0.9}
\renewcommand{\bottomfraction}{0.9}
\renewcommand{\textfraction}{0.05}
\renewcommand{\floatpagefraction}{0.85}
\renewcommand{\dbltopfraction}{0.9}
\setcounter{topnumber}{4}
\setcounter{bottomnumber}{4}
\setcounter{totalnumber}{8}
\setcounter{dbltopnumber}{4}

\section{Extended Results}
\label{sec:supp_results}

The ViT-B task $\times$ ablation matrix (\cref{tab:vitb_matrix}) complements the ViT-S matrix in the main text (\cref{tab:heatmap}). Per-task breakdowns with confidence intervals follow.

\begin{table}[t]
\centering
\caption{ViT-B task $\times$ ablation matrix (layer~11).
CLS: probe top-1 (\%). Corr: correspondence (\%).
Seg: segmentation mIoU (\%). SPair: PCK@0.1 (\%).
$\Delta$: change from Full.}
\label{tab:vitb_matrix}
\scriptsize
\setlength{\tabcolsep}{2pt}
\begin{tabular}{@{}llcccc@{}}
\toprule
& & CLS & Corr & Seg & SPair \\
\midrule
\multirow{2}{*}{DINOv2-B}
  & Full      & 78.7 & 72.9 & 72.0 & 41.2 \\
  & Zero CLS  & 0.1 \dg{-78.6} & 58.9 \dg{-14.0} & 46.1 \dg{-25.9} & 21.3 \dg{-19.9} \\
\midrule
\multirow{3}{*}{\shortstack[l]{DINOv2-B\\+reg}}
  & Full      & 74.5 & 71.2 & 72.3 & 41.1 \\
  & Zero CLS  & 0.1 \dg{-74.4} & 70.4 \dg{-0.8} & 72.3 \dg{0.0} & 41.2 \dg{+0.1} \\
  & Zero Reg  & 55.2 \dg{-19.3} & 63.3 \dg{-7.9} & 64.1 \dg{-8.2} & 28.8 \dg{-12.3} \\
\midrule
\multirow{3}{*}{DINOv3-B}
  & Full      & 73.3 & 77.1 & 83.4 & 37.9 \\
  & Zero CLS  & 0.1 \dg{-73.2} & 79.5 \dg{+2.4} & 82.8 \dg{-0.6} & 37.8 \dg{-0.1} \\
  & Zero Reg  & 36.8 \dg{-36.5} & 61.3 \dg{-15.8} & 59.6 \dg{-23.8} & 19.1 \dg{-18.8} \\
\bottomrule
\end{tabular}
\end{table}

\paragraph{Segmentation.}
\label{sec:supp_segmentation}

\begin{table}[t]
\centering
\caption{Segmentation probe mIoU (\%, ViT-S, VOC2012 val, layer~11,
mean $\pm$ std over 3 seeds).}
\label{tab:segmentation_full}
\small
\begin{tabular}{@{}llc@{}}
\toprule
Model & Ablation & mIoU \\
\midrule
\multirow{2}{*}{DINOv2}
  & Full     & $70.8 \pm 0.10$ \\
  & Zero CLS & $33.7 \pm 0.07$ \\
\midrule
\multirow{3}{*}{DINOv2+reg}
  & Full     & $71.3 \pm 0.07$ \\
  & Zero CLS & $70.7 \pm 0.11$ \\
  & Zero Reg & $61.7 \pm 0.14$ \\
\midrule
\multirow{3}{*}{DINOv3}
  & Full     & $78.5 \pm 0.00$ \\
  & Zero CLS & $78.5 \pm 0.04$ \\
  & Zero Reg & $47.6 \pm 0.08$ \\
\bottomrule
\end{tabular}
\end{table}

\paragraph{Synthetic correspondence.}
\label{sec:supp_correspondence}

\Cref{tab:correspondence} provides per-condition correspondence with bootstrapped CIs.
Register zeroing reduces correspondence from $69$--$79\%$ to $58$--$64\%$ (ViT-S; \cref{tab:correspondence}), visible as collapsed match patterns in \cref{fig:correspondence_qual}.

\begin{table}[b]
\centering
\caption{Patch correspondence (ViT-S, layer~11, 2{,}000 pairs).
GT acc: ground-truth accuracy (\%). Cycle: cycle consistency
(A$\to$B$\to$A). Bootstrapped 95\% CI in brackets.}
\label{tab:correspondence}
\small
\begin{tabular}{@{}llcc@{}}
\toprule
Model & Ablation & GT acc & Cycle \\
\midrule
\multirow{2}{*}{DINOv2}
  & Full     & 72.0 \ci{71.4}{72.7} & 0.468 \\
  & Zero CLS & 56.1 \ci{55.2}{57.1} & 0.411 \\
\midrule
\multirow{3}{*}{DINOv2+reg}
  & Full     & 69.1 \ci{68.4}{69.9} & 0.454 \\
  & Zero CLS & 68.3 \ci{67.6}{69.2} & 0.450 \\
  & Zero Reg & 64.3 \ci{63.4}{65.2} & 0.433 \\
\midrule
\multirow{3}{*}{DINOv3}
  & Full     & 78.9 \ci{78.2}{79.8} & 0.477 \\
  & Zero CLS & 78.2 \ci{77.4}{79.1} & 0.477 \\
  & Zero Reg & 57.8 \ci{56.5}{59.2} & 0.331 \\
\bottomrule
\end{tabular}
\end{table}

\paragraph{Tolerance sensitivity.}
\label{sec:supp_tolerance}
Ablation patterns are stable across tolerance thresholds (\cref{tab:tolerance}).
DINOv3 shows the largest register-zeroing drop at all thresholds ($-18.2$ to $-21.1$\,pp).

\begin{table}[t]
\centering
\caption{Correspondence accuracy (\%) at different tolerance
thresholds (ViT-S, layer~11). Tol\,=\,0: exact match; 1: $\pm$1 patch
(default); 2: $\pm$2 patches.}
\label{tab:tolerance}
\small
\setlength{\tabcolsep}{3pt}
\begin{tabular}{@{}llccc@{}}
\toprule
Model & Ablation & Tol\,=\,0 & Tol\,=\,1 & Tol\,=\,2 \\
\midrule
\multirow{2}{*}{DINOv2}
  & Full     & 38.5 & 72.0 & 80.3 \\
  & Zero CLS & 28.0 & 56.1 & 65.2 \\
\midrule
\multirow{3}{*}{\shortstack[l]{DINOv2\\+reg}}
  & Full     & 37.5 & 69.1 & 77.1 \\
  & Zero CLS & 36.8 & 68.3 & 76.4 \\
  & Zero Reg & 34.0 & 64.3 & 72.7 \\
\midrule
\multirow{3}{*}{DINOv3}
  & Full     & 45.2 & 78.9 & 85.3 \\
  & Zero CLS & 45.0 & 78.2 & 84.0 \\
  & Zero Reg & 27.0 & 57.8 & 66.4 \\
\bottomrule
\end{tabular}
\end{table}

\paragraph{SPair-71k quantization ceiling.}
\label{sec:supp_spair}
DINOv3 uses $16$-pixel patches ($14{\times}14$ grid, $196$ tokens) vs.\ DINOv2's $14$-pixel patches ($16{\times}16$ grid, $256$ tokens).
An oracle quantization test yields ceilings of $97.9\%$ ($16{\times}16$ grid) and $96.5\%$ ($14{\times}14$ grid)---a $1.4$\,pp gap that accounts for only a fraction of the $6.6$\,pp absolute PCK difference, so grid resolution is not the dominant factor.
ViT-B SPair results follow the same asymmetry (\cref{tab:vitb_matrix}).

\paragraph{SPair-71k correspondence.}
\label{sec:supp_spair_results}
The SPair-71k results table is in the main text (\cref{tab:spair}).

\section{Controls and Statistical Tests}
\label{sec:supp_controls}

\paragraph{Random-patch negative control.}
\label{sec:supp_random_patch}
Zeroing 4 random patch tokens (5 seeds) causes $\leq\!1$\,pp CLS drop for ViT-S and $\leq\!2.3$\,pp for ViT-B (\cref{tab:random_patch}), vs.\ $-18.9$ / $-36.6$\,pp for register zeroing.

\begin{table}[b]
\centering
\caption{Random-patch negative control: CLS top-1 (\%) when zeroing 4
random patch tokens (ViT-S and ViT-B, layer~11, 5 seeds).
$\Delta$: change from Full baseline (\cref{tab:heatmap,tab:vitb_matrix}).
Reg-zero $\Delta$ reproduced for comparison.}
\label{tab:random_patch}
\scriptsize
\setlength{\tabcolsep}{3pt}
\begin{tabular}{@{}llccc@{}}
\toprule
Model & Condition & CLS (\%) & $\Delta$ & Reg-zero $\Delta$ \\
\midrule
DINOv2       & Rand.\ patch  & $72.4 \pm 0.10$ & $-0.8$ & --- \\
DINOv2+reg   & Rand.\ patch  & $66.3 \pm 0.03$ & $-1.0$ & $-18.9$ \\
DINOv3       & Rand.\ patch  & $61.5 \pm 0.05$ & $-0.5$ & $-36.6$ \\
\midrule
DINOv2-B     & Rand.\ patch  & $78.4 \pm 0.07$ & $-0.3$ & --- \\
DINOv2-B+reg & Rand.\ patch  & $74.5 \pm 0.04$ & $\phantom{-}0.0$ & $-19.3$ \\
DINOv3-B     & Rand.\ patch  & $71.0 \pm 0.06$ & $-2.3$ & $-36.5$ \\
\bottomrule
\end{tabular}
\end{table}

\paragraph{Mean-substitution control.}
\label{sec:supp_meansub}
Replacing registers with per-layer dataset-mean activations (5{,}000 images; \cref{tab:controls} in main text) has negligible effect ($-0.3$\,pp and $+0.1$\,pp) vs.\ $-18.9$ and $-36.6$\,pp under zeroing. Mean-substituting CLS yields $0.1\%$ (same as zeroing), indicating CLS carries all image-specific class signal.
Classification accuracy under mean-substitution is insensitive to calibration set size: varying $N$ from 100 to 5{,}000 images changes accuracy by ${\leq}0.1$\,pp for both DINOv2+reg ($67.0$--$67.1\%$) and DINOv3 ($62.0$--$62.1\%$), indicating the control does not depend on precise mean estimates.

\paragraph{Noise-substitution and register-shuffling controls.}
\label{sec:supp_noise_shuffle}
\emph{Noise-substitution} replaces register outputs at each layer with Gaussian noise matched in per-dimension mean and variance (calibrated on 5{,}000 images), preserving marginal statistics but destroying register-specific structure.
\emph{Register shuffling} permutes register activations across images within each batch (independently at each layer), preserving real activation structure but breaking image-specific routing.
All three controls preserve performance across classification, correspondence, and segmentation (see \cref{tab:controls} in main text for complete results).

\paragraph{Statistical significance.}
\label{sec:supp_pvalues}
Per-image or per-token outcome differences are tested with sign-flip permutation tests (10{,}000 permutations; \cref{tab:pvalues}).
The two dissociation comparisons are significant at $p < 0.001$; the scale comparison ($p = 0.80$) confirms register dependence is consistent across ViT-S and ViT-B.

\begin{table}[t]
\centering
\caption{Paired permutation test p-values for headline comparisons
(10{,}000 permutations, two-sided).}
\label{tab:pvalues}
\scriptsize
\setlength{\tabcolsep}{3pt}
\begin{tabular}{@{}p{5.5cm}ccc@{}}
\toprule
\textbf{Comparison} & \textbf{Observed $\Delta$} & \textbf{$p$} \\
\midrule
DINOv3 vs.\ DINOv2+reg $\Delta_\text{ZeroReg}$ (CLS) &
  $17.7$\,pp & $< 0.001$ \\
CLS-zeroing buffering (Seg, with vs.\ without regs) &
  $10.7$\,pp & $< 0.001$ \\
ViT-S vs.\ ViT-B register dependence (DINOv3 $\Delta_\text{ZeroReg}$) &
  $0.1$\,pp & $0.80$ (ns) \\
\bottomrule
\end{tabular}
\end{table}

\section{Representation Geometry}
\label{sec:supp_gram}

\begin{figure}[t]
  \centering
  \includegraphics[width=\columnwidth]{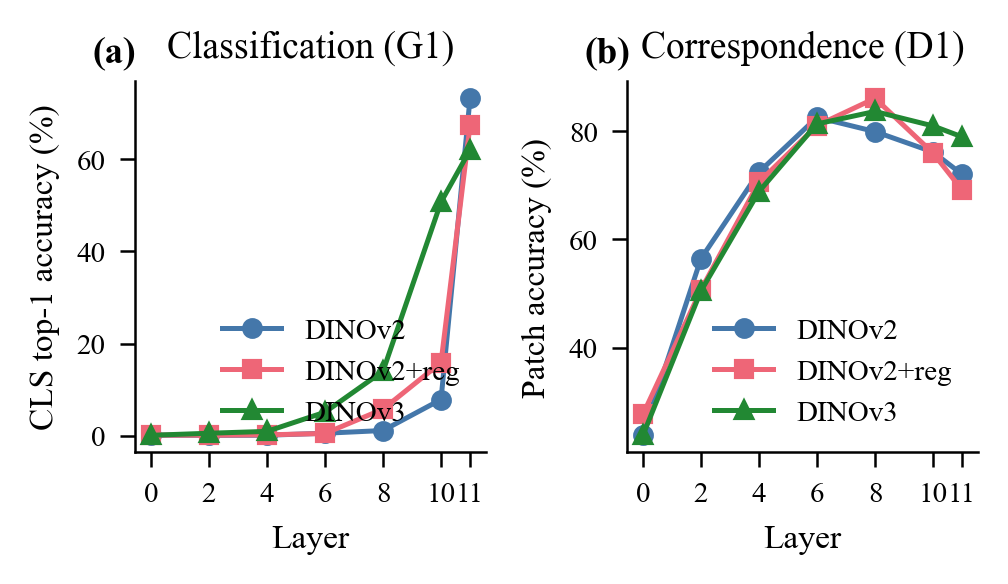}
  \caption{\textbf{Task performance across transformer layers.}
  \textbf{(a)}~CLS classification (linear probe, 50 epochs, 1 seed).
  Classification emerges at layers~10--11.
  \textbf{(b)}~Patch correspondence (tolerance = 1).
  Correspondence peaks at layers~6--8 then declines, except DINOv3
  which maintains $78.9\%$ at layer~11. ViT-S models.}
  \label{fig:layer_sweep}
\end{figure}

Effective rank across layers reveals when patch compression and register dependence emerge (\cref{tab:gram_layers,tab:gram_full,fig:layer_sweep}; see also \cref{fig:gram_dependence}).
DINOv3 is already compressed at layer~6 (effective rank $6.4$ vs.\ $32.3$ for DINOv2), yet register dependence for classification emerges only at layers~10--11.
In DINOv3, register zeroing \emph{improves} CLS accuracy at intermediate layers before becoming strongly detrimental at the final layers, suggesting registers become structurally expected only as classification information consolidates.
The compression trajectory also differs across models: DINOv2+reg has higher effective rank than DINOv2 at layer~6 ($41.7$ vs.\ $32.3$) before dropping below it at layer~11 ($8.7$ vs.\ $13.5$), indicating that register-induced compression is not uniform across layers.

\begin{table}[b]
\centering
\caption{Effective rank (median $\pm$ std) at layers~6 and~11
(ViT-S, 500 ImageNet images).
Layer~11 values match the Full rows of \cref{tab:gram_full}.}
\label{tab:gram_layers}
\small
\begin{tabular}{@{}lcc@{}}
\toprule
Model & Layer 6 & Layer 11 \\
\midrule
DINOv2     & $32.3 \pm 6.3$  & $13.5 \pm 4.5$ \\
DINOv2+reg & $41.7 \pm 7.4$  & $\phantom{0}8.7 \pm 3.3$ \\
DINOv3     & $\phantom{0}6.4 \pm 7.0$  & $\phantom{0}4.0 \pm 1.2$ \\
\bottomrule
\end{tabular}
\end{table}

\begin{table}[t]
\centering
\caption{Median Gram statistics (ViT-S, layer~11, 2{,}000 images).
DINOv3 entropy omitted: its $14{\times}14$ patch grid yields a
$196{\times}196$ Gram matrix vs.\ $256{\times}256$ for DINOv2 models,
making eigenspectrum entropy incomparable.}
\label{tab:gram_full}
\small
\begin{tabular}{@{}llcc@{}}
\toprule
Model & Ablation & Erank & Entropy \\
\midrule
\multirow{2}{*}{DINOv2}
  & Full     & 13.5 & 2.61 \\
  & Zero CLS & 18.1 & 2.90 \\
\midrule
\multirow{3}{*}{DINOv2+reg}
  & Full     & \phantom{0}8.7 & 2.16 \\
  & Zero CLS & \phantom{0}9.8 & 2.29 \\
  & Zero Reg & 11.3 & 2.43 \\
\midrule
\multirow{3}{*}{DINOv3}
  & Full     & \phantom{0}4.0 & --- \\
  & Zero CLS & \phantom{0}4.5 & --- \\
  & Zero Reg & \phantom{0}5.1 & --- \\
\bottomrule
\end{tabular}
\end{table}

\section{Mechanistic Analysis}
\label{sec:supp_mechanistic}

\paragraph{Attention flow across layers.}
\cref{fig:attention_flow} traces how attention mass distributes between token types at each of the 12 transformer layers.
In DINOv2 (no registers), CLS self-attention dominates early layers then declines.
In both register models, register attention share builds \emph{gradually} from mid-layers: DINOv2+reg stabilizes at ${\sim}20\%$ CLS$\to$registers by layer~6, while DINOv3 ramps more steeply to $28.7\%$ at layer~11.
This gradual ramp contrasts with the abrupt emergence of classification accuracy at layers~10--11, dissociating attention routing from functional dependence.

\paragraph{Attention pattern analysis.}
\cref{fig:rewiring} in the main text compares attention patterns under register zeroing vs.\ mean-substitution across all 12 layers.
At ViT-S scale, register zeroing yields last-layer JS divergence of $0.144$ (DINOv2+reg) and $0.177$ (DINOv3), while mean-substitution yields $0.005$ and $0.001$ respectively---a $29{\times}$ and $177{\times}$ gap respectively (using the rounded values reported here).
Divergence is identically zero at layer~0 (same input) and amplifies across layers, consistent with the distributional shift from zero vectors cascading rather than remaining a single-layer effect.
CLS zeroing produces much smaller divergence ($0.026$ and $0.010$), consistent with CLS being a downstream reader rather than a structural element whose removal causes large distributional shift.

\textbf{Per-register dose-response.}
The register whose removal causes the greatest attention disruption matches the register with the highest decodable class information: DINOv2+reg R2 (JS\,$=$\,$0.062$, vs.\ $0.020$--$0.025$ for other registers) and DINOv3 R3 (JS\,$=$\,$0.076$, vs.\ $0.006$--$0.022$).
Note that zeroing individual registers is also a distribution-shifting intervention, so these results should be interpreted as measuring sensitivity to distributional shift rather than functional dependence on register content.

\textbf{Scale consistency.}
At ViT-B scale, the pattern replicates: register zeroing yields last-layer JS of $0.232$ (DINOv2-B+reg) and $0.183$ (DINOv3-B), while mean-substitution yields $0.009$ and $0.002$ (\cref{fig:rewiring}a, lighter lines).

The attention-flow and PCA-projection figures are in the main text (\cref{fig:attention_flow,fig:pca_ablation}).

\begin{figure}[t]
  \centering
  \includegraphics[width=0.95\columnwidth]{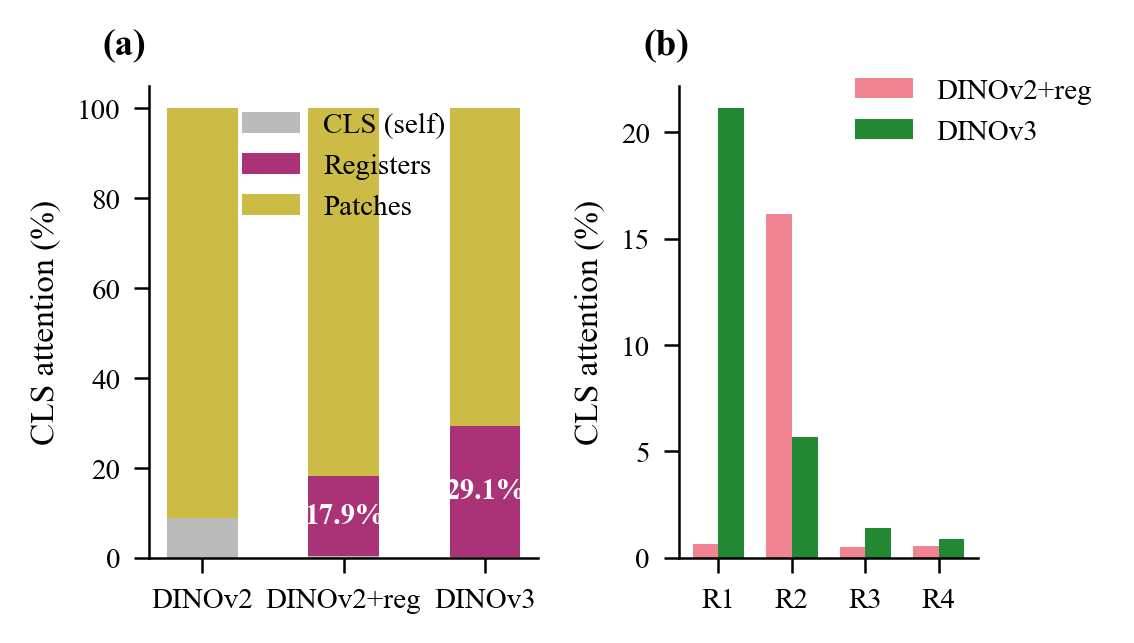}
  \caption{\textbf{CLS attention (ViT-S, last layer, 200 images).}
  \textbf{(a)}~CLS attention fraction per token type.
  DINOv2+reg: $17.9\%$ to registers; DINOv3: $29.1\%$.
  \textbf{(b)}~Per-register breakdown.
  This routing structure is maintained under all plausible register replacements.}
  \label{fig:attention}
\end{figure}

\paragraph{Register token analysis.}
\label{sec:supp_registers}
Initial per-register lesions and probes suggested register specialization.
However, the substitution controls in \cref{tab:controls} show that these patterns need not reflect functional dependence: class information is decodable from individual registers, yet models do not require it for any measured downstream task.
We report these analyses as exploratory and descriptive; they characterize representational structure but not functional necessity.

\section{ViT-B Scale Validation}
\label{sec:supp_vitb}

We replicate the zero-ablation experiments at ViT-B scale: DINOv2-B (86.6M params),
DINOv2-B+reg (with four register tokens), and
DINOv3-B (85.7M params).
The full ViT-B ablation matrix is in \cref{tab:vitb_matrix};
replacement controls are in the main text (\cref{tab:vitb_controls}).

\paragraph{Task-level replication.}
ViT-B absolute accuracies are higher but ablation deltas are nearly identical to ViT-S.
Classification: DINOv3-B loses $-36.5$\,pp \ci{-38.2}{-34.6} under register zeroing (vs.\ $-36.6$ at ViT-S); DINOv2-B+reg loses $-19.3$\,pp \ci{-21.2}{-17.5} (vs.\ $-18.9$).
Correspondence: CLS zeroing degrades DINOv2-B by $-14.0$\,pp but is negligible with registers; register zeroing hits DINOv3-B hardest ($-15.8$ vs.\ $-7.9$\,pp).
SPair-71k: same pattern ($-18.8$\,pp DINOv3-B vs.\ $-12.3$\,pp DINOv2-B+reg).
Segmentation: CLS zeroing drops DINOv2-B by $-25.9$\,pp but $\leq\!0.6$\,pp with registers; register zeroing again hits DINOv3-B hardest ($-23.8$ vs.\ $-8.2$\,pp).
The Gram compression pattern replicates: effective rank $20.1 \to 13.9 \to 6.6$ ($3.0{\times}$ compression from DINOv2-B to DINOv3-B, vs.\ $3.4{\times}$ at ViT-S).

\paragraph{Attention and per-register patterns.}
ViT-B attention routing mirrors ViT-S: DINOv2-B+reg routes $25.6\%$
of CLS attention to registers (R4 dominant at $23.5\%$);
DINOv3-B routes $30.6\%$ (R1 largest at $15.4\%$, more distributed).
Per-register zero-lesions show a distributed pattern in DINOv3-B
(R3: $-1.7$\,pp, R2: $-1.5$\,pp, R1: $-1.1$\,pp, R4: $-0.9$\,pp)
while DINOv2-B+reg shows minimal individual effects ($\leq 1.1$\,pp);
however, zeroing individual registers is also a distribution-shifting intervention (\cref{sec:supp_registers}).
The identity of the dominant attention register shifts from R2 (ViT-S) to R4 (ViT-B), suggesting that register differentiation is not fixed across scales but emerges from training dynamics.

\section{Experimental Details}
\label{sec:supp_details}

\paragraph{Feature extraction.}
All features are extracted using HuggingFace
\texttt{transformers} (\texttt{facebook/dinov2-small} and
\texttt{facebook/dinov2-with-registers-small} for ViT-S;
\texttt{facebook/dinov2-base} and
\texttt{facebook/dinov2-with-registers-base} for ViT-B).
DINOv3 models are loaded via \texttt{torch.hub} with locally cached weights (\texttt{dinov3\_vits16} and \texttt{dinov3\_vitb16}).
Input images are resized to $224 \times 224$ and normalized with
ImageNet statistics.
We extract features from the final block output after the model's terminal LayerNorm (layer~11 of 12).

\paragraph{Ablation hooks.}
PyTorch forward hooks replace the relevant token positions after every block output, starting from block~1.
Zero-ablation substitutes $\mathbf{0} \in \mathbb{R}^{d}$; mean-substitution uses per-layer dataset-mean activations calibrated on 5{,}000 images; noise-substitution uses per-layer Gaussian noise matched in mean and variance; register shuffling permutes activations across images within each batch independently at each layer.

\paragraph{Linear probe training.}
Single linear layer ($d \to K$).
SGD with momentum 0.9, weight decay $0.1$, learning rate 0.01 with cosine annealing, 100 epochs, batch size 256.
Stratified 80/20 split (40{,}000 / 10{,}000 images, seed 42).

\paragraph{Segmentation probe training.}
Single linear layer ($d \to 21$) per patch token;
masks downsampled to the patch grid via nearest-neighbor interpolation.
AdamW, weight decay $10^{-2}$, learning rate $10^{-3}$, constant, 100 epochs.
Per-pixel cross-entropy, ignoring void (index 255).

\paragraph{kNN retrieval.}
2{,}000 ImageNet val images, each producing two augmented views
(\texttt{RandomResizedCrop}, \texttt{ColorJitter}, \texttt{RandomHorizontalFlip}).
Cosine similarity; R@1 with 1{,}000 bootstrap resamples.

\paragraph{Patch correspondence.}
Same augmentation pipeline as kNN, with crop coordinates recorded for ground-truth spatial correspondence.
A match is correct if the nearest-neighbor patch falls within 1 patch of ground truth; cycle consistency requires the chain A$\to$B$\to$A to return to the exact original patch.

\paragraph{SPair-71k evaluation.}
Images resized to $224 \times 224$; source keypoints mapped to the patch grid; correspondence predicted via cosine similarity.
PCK@0.1: correct if Euclidean distance $< 0.1 \times \max(h_\text{bbox}, w_\text{bbox})$.
Only mutually visible keypoints evaluated; 1{,}000 bootstrap resamples.

\paragraph{Attention analysis.}
Attention weights from the final block (\texttt{attn\_implementation="eager"} for HF models).
CLS row averaged over heads, then over 200 images.
Layer-sweep probes use 50 epochs / 1 seed (lower absolute accuracy than main-text probes; relative layer-wise patterns are stable).

\paragraph{Attention rewiring analysis.}
Jensen--Shannon divergence is computed between full and ablated attention distributions at every layer and head (1{,}000 images, all 12 layers).
Per-head JS values are averaged across heads and images to produce per-layer divergence curves.

\paragraph{Compute.}
Single NVIDIA RTX 4090 (24\,GB); total GPU time $\approx$12--15 hours (feature extraction across 6 models $\times$ multiple ablation conditions $\times$ 4 datasets, probe training, attention analysis, and mean-activation calibration).

\end{document}